\documentclass[10pt,twocolumn,letterpaper]{article}

\usepackage{iccv}
\usepackage{times}
\usepackage{epsfig}
\usepackage{graphicx}
\usepackage{amsmath}
\usepackage{amssymb}

\usepackage[accsupp]{axessibility}

\usepackage{subcaption}
\usepackage{multirow}
\usepackage{bm}
\usepackage{comment}
\usepackage{multirow}
\usepackage[misc]{ifsym}

\usepackage[breaklinks=true,bookmarks=false, hidelinks]{hyperref}

\iccvfinalcopy 

\def\iccvPaperID{9628} 

\ificcvfinal\fi

\usepackage{siunitx}
\usepackage{booktabs, makecell}
\usepackage{subcaption}
\usepackage{threeparttable}
\usepackage{mathtools}

\newcommand{\sota}{state-of-the-art}

\newcommand{\rom}[1]{\uppercase\expandafter{\romannumeral #1\relax}}
\newcommand{\romsm}[1]{\lowercase\expandafter{\romannumeral #1\relax}}
\renewcommand{\eqref}[1]{Eq.\,(\ref{#1})}

\begin{document}

\title{MMST-ViT: Climate Change-aware Crop Yield Prediction via Multi-Modal Spatial-Temporal Vision Transformer}

\author{
   Fudong Lin$^{1}$, ~Summer Crawford$^{2}$, ~Kaleb Guillot$^{2}$,   ~Yihe Zhang$^{2}$,  ~Yan Chen$^{3}$, 
   ~Xu Yuan$^{1}$\thanks{Corresponding author: Dr. Xu Yuan (xyuan@udel.edu)}~,   \\
   Li Chen$^{2}$,   ~Shelby Williams$^{2}$,  ~Robert Minvielle$^{2}$,      ~Xiangming Xiao$^{4}$,  Drew Gholson$^{5}$, Nicolas Ashwell$^{5}$, \\
    ~Tri Setiyono$^{6}$,   ~Brenda Tubana$^{7}$, ~Lu Peng$^{8}$, Magdy Bayoumi$^{2}$, ~Nian-Feng Tzeng$^{2}$ \\
   $^{1}$ University of Delaware, $^{2}$ University of Louisiana at Lafayette, $^{3}$ University of Connecticut, \\
   $^{4}$ University of Oklahoma, $^{5}$ Mississippi State University, $^{6}$ Louisiana State University, \\
   $^{7}$ LSU AgCenter, $^{8}$ Tulane University   
}

\maketitle
\ificcvfinal\thispagestyle{empty}\fi

\begin{abstract}
   Precise crop yield prediction provides valuable information 
   for agricultural planning and decision-making processes. 
   However, timely predicting crop yields remains challenging
   as crop growth is sensitive to growing season weather variation and climate change.
   In this work, we develop a deep learning-based solution,
   namely \underline{M}ulti-\underline{M}odal \underline{S}patial-\underline{T}emporal \underline{Vi}sion \underline{T}ransformer~(MMST-ViT), 
   for predicting crop yields at the county level across the United States,
   by considering the effects of short-term meteorological variations during the growing season
   and the long-term climate change on crops.
   Specifically, our MMST-ViT consists of a Multi-Modal Transformer, a Spatial Transformer, and a Temporal Transformer.
   The Multi-Modal Transformer leverages both visual remote sensing data 
   and short-term meteorological data for modeling the effect of growing season weather variations on crop growth.
   The Spatial Transformer learns the high-resolution spatial dependency among counties for accurate agricultural tracking.
   The Temporal Transformer captures the long-range temporal dependency 
   for learning the impact of long-term climate change on crops.
   Meanwhile, we also devise a novel multi-modal contrastive learning technique
   to pre-train our model without extensive human supervision. 
   Hence, our MMST-ViT captures the impacts of both short-term weather variations and long-term climate change on crops
   by leveraging both satellite images and meteorological data.
   We have conducted extensive experiments on over $200$ counties in the United States, 
   with the experimental results exhibiting that our MMST-ViT outperforms its counterparts under three performance metrics of interest.
   Our dataset and code are available at \textcolor{magenta}{\url{https://github.com/fudong03/MMST-ViT}}.
\end{abstract}


\section{Introduction}\label{sec:intro}
Accurate crop yield prediction is essential for
agricultural planning and advisory processes~\cite{khaki2021simultaneous}, 
informed economic decisions~\cite{ansarifar2021interaction},
and global food security~\cite{mourtzinis2021advancing}.
However, predicting crop yields precisely is challenging
as it requires to consider the effects of 
\romsm{1}) short-term weather variations, 
governed by the meteorological data during the growing season,
and \romsm{2}) long-term climate change, 
governed by historical meteorological factors, on crops simultaneously.
Meanwhile, precise crop tracking relies on high-resolution remote sensing data.
While process-based prediction approaches 
\cite{tao2009modelling,chenu2017contribution,jones2017toward,xu2019global} exist,
they often suffer from high inaccuracies due to their strong assumptions on management practices~\cite{fan:aaai23:crop_prediction}. 
On the other hand, 
motivated by the recent success of deep neural networks
\cite{alex:nips12:alex_net,he:cvpr16:resnet,vaswani:nips17:attention,dosovitskiy:iclr21:vit,
fudong:ijcai21:oversampling,yihe:ecml-pkdd21:storm,radford:icml21:clip,fudong:cikm22:cascade_vae,he:cvpr22:mae,fudong:ecml23:storm,lai:iccv23:pad_clip,chen2023learning},
deep learning (DL)-based methods have been widely adopted for crop yield predictions,
due to their effectiveness in accurate agricultural tracking~\cite{khaki2019crop,khaki2021simultaneous}
and their potent capabilities in capturing the spatial and temporal variation of meteorological data~\cite{khaki2020cnn,fan:aaai23:crop_prediction}.

So far, DL-based solutions for crop yield predictions 
can be roughly grouped into two categories, 
\ie, remote sensing data-based and meteorological data-based approaches.
The former~\cite{khaki2019crop,khaki2021simultaneous,
wu2021spatiotemporal,falco2021influence,garnot:iccv21:panoptic,tseng2021cropharvest,cheng2022high}
employs such remote sensing data as satellite images, unmanned aerial vehicle~(UAV)-based imagery data,
or vegetation indices to estimate the annual crop yield,
while the latter~\cite{gandhi2016rice,akhavizadegan2021time,mourtzinis2021advancing,shahhosseini2021coupling,turchetta:nips22:cycle_gym}
predicts the crop yield by using meteorological parameter data, 
including temperature, precipitation, vapor pressure deficit, \etc
However, the former overlooks the direct impact of meteorological parameters on crop growth,
while the latter lacks high-resolution remote sensing data for accurate agricultural tracking.

A recent study~\cite{rolnick:survey23:climate_change} has reported 
that the long-term climate change would gradually decrease the crop yield. 
Driven by this discovery,
follow-up pursuits have attempted to explore the effect of long-term climate change on crops.
For example, the CNN-RNN model~\cite{khaki2020cnn} 
demonstrates that the crop yield prediction
can benefit from historical meteorological data, 
which is essential for measuring the impacts of climate change.
Later, GNN-RNN~\cite{fan:aaai23:crop_prediction} extends CNN-RNN
by framing the crop yield prediction as the Spatial-Temporal forecasting problem.
It employs Graph Neural Networks~(GNN) and Long Short-Term Memory~(LSTM)~\cite{hochreiter1997lstm}
respectively for learning spatial dependency among neighborhood counties
and for capturing the impact of long-term meteorological data on crops.
However, both of them only take into account the meteorological data for predictions, 
failing to leverage the remote sensing data for accurate agricultural tracking.

In this work, we aim to develop a new DL-based solution 
for predicting crop yields at the county level across the United States, 
by using both visual remote sensing data (from the Sentinel-2 satellite imagery~\cite{sentinel-hub})
and numerical meteorological data (from the HRRR model~\cite{hrrr}). 
Our solution has two main goals. 
First, it captures the impacts of both short-term growing season weather variations 
and long-term climate change on crops.
Second, it aims to leverage high-resolution remote sensing data for accurate agricultural tracking.
To achieve our goals,
we propose the Multi-Modal Spatial-Temporal Vision Transformer~(MMST-ViT),
motivated by the recent success of Vision Transformers (ViT)~\cite{dosovitskiy:iclr21:vit}.
To the best of our knowledge,
MMST-ViT is the first ViT-based model for real-world crop yield prediction.
It advances previous CNN/GNN-based and RNN-based counterparts 
respectively with better generalization to the multi-model data 
and with more powerful abilities in capturing long-term temporal dependency.
Its three components of a Multi-Modal Transformer, a Spatial Transformer, and a Temporal Transformer 
are each equipped with one novel Multi-Head Attention~(MHA) mechanism~\cite{vaswani:nips17:attention}. 
Specifically, the Multi-Modal Transformer leverages satellite images and meteorological data during the growing season
for capturing the direct impact of short-term weather variations on crop growth.
The Spatial Transformer learns high-resolution spatial dependency among counties for precise crop tracking.
The Temporal Transformer captures the effects of long-term climate change on crops.
Since ViT-based models are prone to overfitting~\cite{dosovitskiy:iclr21:vit},
we also develop a novel multi-modal contrastive learning technique
that pre-trains our Multi-Modal Transformer without requiring human supervision.
We have conducted experiments on over $200$ counties in the United States.
The experimental results exhibit that our MMST-ViT outperforms its 
state-of-the-art counterparts under three performance metrics of interest.
For example, on the soybean prediction, our MMST-ViT achieves the lowest Root Mean Square Error (RMSE) value
of $3.9$, the highest R-squared (R$^{2}$) value of $0.843$, and the best Pearson Correlation Coefficient (Corr) value of $0.918$.

\section{Related Work}\label{sec:rw}

\par\smallskip\noindent
{\bf Vision Transformers.} 
Adopted from Transformers~\cite{vaswani:nips17:attention} in natural language processing, 
Vision Transformers~(ViT) have 
exhibited commendable performance in various computer vision tasks. 
The original ViT~\cite{dosovitskiy:iclr21:vit} first partitions an image into a set of image patches, 
then applies Multi-Head Attention~(MHA)~\cite{vaswani:nips17:attention} over the patches, 
and finally utilizes a learnable classification token to capture global visual representation for image classification. 
Several subsequent methods have been developed, 
including DeiT~\cite{touvron:icml21:deit} for data-efficient ViT through knowledge distillation, 
Swin~\cite{liu:iccv21:swin} for computation-efficient ViT using shifted windows, 
PVT~\cite{wang:iccv21:pvt} for dense prediction tasks, 
and MAE~\cite{he:cvpr22:mae} for self-supervised learning, 
among others~\cite{caron:iccv21:dino,heo:iccv21:pit,fan:iccv21:mvit,yuan:iccv21:t2tvit,arnab:iccv21:vivit,wang:cvm22:pvtv2,li:cvpr22:mvit2,tong2022videomae}.
However, applying prior ViT approaches to real-world crop yield prediction is challenging
due to its needs of addressing the multi-modal inputs, of learning high-resolution spatial dependency, and of capturing long-range temporal dependency.
Our work, based on ViT, advances existing methods by proposing three novel Multi-Head Attention (MHA) mechanisms
respectively for leveraging both visual remote sensing data and numerical meteorological data,
learning global spatial representation from multiple high-resolution data,
and capturing the global temporal representation for measuring the long-term climate change effect.
Additionally, we develop a new multi-modal contrastive learning technique 
to pre-train our multi-modal model for better prediction performance.

\par\smallskip\noindent
{\bf Deep Learning (DL) for Crop Yield Prediction.} 
DL has been widely adopted for real-world crop yield predictions.
Such prediction studies can be grouped into two categories: remote sensing data-based 
and meteorological data-based approaches.
The former uses satellite images, unmanned aerial vehicle (UAV) data, 
or vegetation indices to estimate crop yields.
Its prominent studies include the use of UAV-acquired RGB images~\cite{khaki2019crop} 
for predicting in-season crop yields,
YieldNet~\cite{khaki2021simultaneous} which resorts to transfer learning 
for simultaneously estimating the yields of multiple crop types,
among many others~\cite{wu2021spatiotemporal,falco2021influence,garnot:iccv21:panoptic,tseng2021cropharvest,cheng2022high}.
By contrast, the latter utilizes deep neural networks (DNNs) 
to capture the impact of meteorological parameters on crop yields,
including CNN-RNN~\cite{khaki2020cnn} which incorporates long-term meteorological data, 
GNN-RNN~\cite{fan:aaai23:crop_prediction} which extends CNN-RNN
by using Graph Neural Networks (GNN) for learning spatial information,
and others~\cite{gandhi2016rice,akhavizadegan2021time,mourtzinis2021advancing,shahhosseini2021coupling,turchetta:nips22:cycle_gym}.
However, the remote sensing data-based solutions overlook the impact of meteorological parameters on crops, 
while the meteorological data-based solutions often fail to incorporate the remote sensing data, 
known to be crucial for accurate agricultural tracking.
Our work differs from prior studies in two aspects. 
First, it leverages both remote sensing data and meteorological data 
for capturing the impacts of both short-term growing season meteorological variations
and the long-term climate change on crops.
Second, 
it is the first to use Vision Transformers for crop yield predictions, 
advancing previous CNN/GNN- and RNN-based models respectively with better generalization to multi-modal data  
and with higher abilities for capturing long ranges of temporal dependency.

\section{Datasets}
\label{sec:pre}

In this study, we utilize three types of data for accurate county-level crop yield predictions: 
\romsm{1}) crop data from the United States Department of Agriculture (USDA), 
\romsm{2}) meteorological data from the High-Resolution Rapid Refresh (HRRR), 
and \romsm{3}) remote sensing data from the Sentinel-2 satellite, as outlined below.

\par\smallskip\noindent
{\bf USDA Crop Dataset.} 
The dataset, sourced from the United States Department of Agriculture (USDA)~\cite{usda}, 
provides annual crop data for major crops grown in the United States~(U.S.), 
including corn, cotton, soybean, winter wheat, \etc, on a county-level basis. 
It covers crop information such as production and yield from $2017$ to $2022$
(as listed in the second row of Table~\ref{tab:exp-dataset-overview}).

\par\smallskip\noindent
{\bf HRRR Computed Dataset.} 
The dataset, obtained from the High-Resolution Rapid Refresh atmospheric model (HRRR)~\cite{hrrr},
provides high-resolution meteorological data for the contiguous U.S. continent.
It covers $9$ weather parameters from $2017$ to $2022$
(see the third row in Table~\ref{tab:exp-dataset-overview} for more information).

\par
\begin{figure} [!t] 
    \captionsetup[subfigure]{justification=centering}
    \begin{subfigure}[t]{0.24\textwidth}
        \centering
        \includegraphics[width=\textwidth]{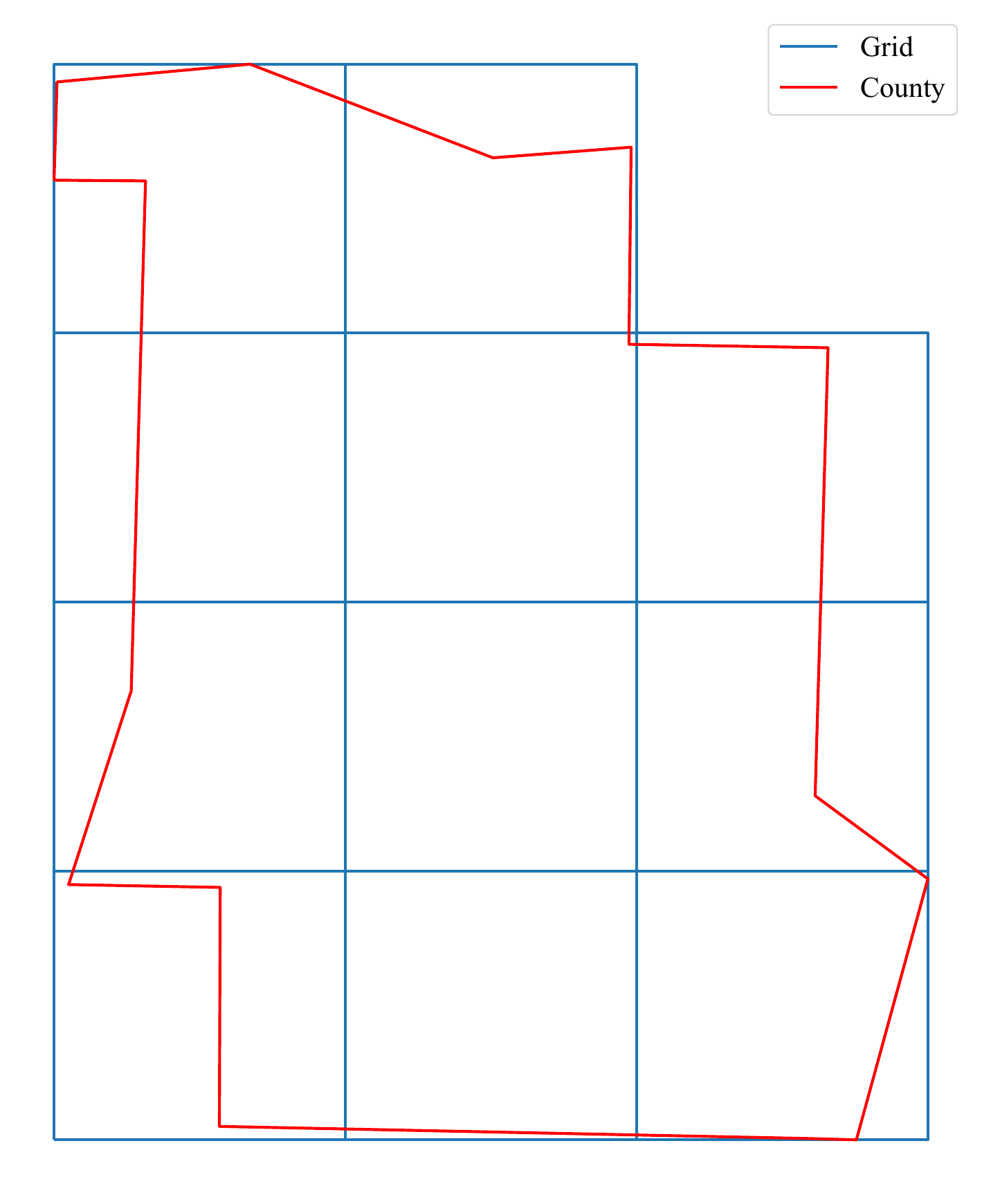}
        \caption{Grid Example}
        \label{fig:dataset-county-grid}
    \end{subfigure}
    \begin{subfigure}[t]{0.22\textwidth}
        \centering
        \includegraphics[width=\textwidth]{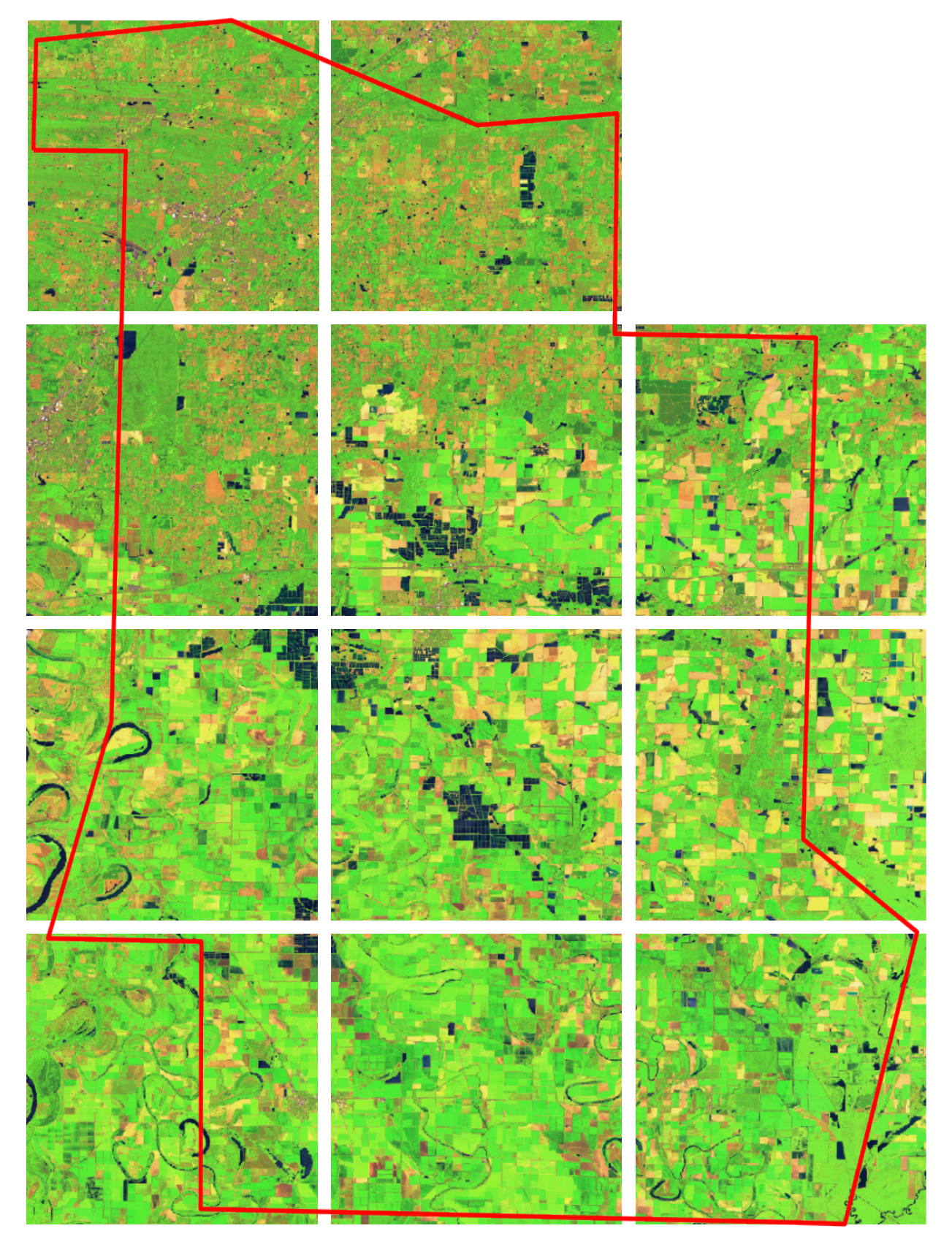}
        \caption{Sentinel-2 Imagery}
        \label{fig:dataset-county-image-mapping}
    \end{subfigure}
    \vspace{-0.5em}
    \caption{
    Illustration of partitioning a county into multiple grids at the $9$km high-resolution.
    (a) An example of county partitioning,
    with the red line segments indicating the geometry boundary for the county and the blue line segments representing the high-resolution grids.
    (b) The resulting satellite images  in the Sentinel-2 Imagery,
    with each composed of $384 \rm{\times} 384$ pixels, depicting an area of $9 \rm{\times} 9$ km.
    }
    \label{fig:dataset-partition}
    \vspace{-1.5 em}
\end{figure}

\smallskip\noindent
{\bf Sentinel-2 Imagery.} 
Sentinel-2 imagery is a set of images 
captured by the Sentinel-2 Earth observation satellite. 
It provides agriculture imagery for the contiguous U.S. continent
from $2017$ to $2022$ with a 2-week interval. 
Since precise agricultural tracking requires high-resolution remote sensing data,
the image of a county is partitioned into multiple fine-grained grids ($9 \rm{\times} 9$ km).
Figure~\ref{fig:dataset-partition} shows an example of county partitioning.

\begin{table}[!t]
    \scriptsize
  \centering
  \setlength\tabcolsep{5 pt}
  \caption{ Overview of USDA Crop Dataset and HRRR Computed Dataset
      }
  \vspace{-0.5 em}
  \begin{tabular}{@{}|c|c|@{}}
    \toprule
    Dataset  & Parameters                                                                                                                                                                                                         \\ \midrule
    USDA     & Production, Yield   \\ \midrule
    HRRR     & \begin{tabular}[c]{@{}c@{}} Averaged Temperature, Maximal Temperature, Minimal Temperature, \\ Precipitation,  Relative Humidity, Wind Gust, Wind Speed, \\ Downward Shortwave Radiation Flux, Vapor Pressure Deficit\end{tabular}    \\ \bottomrule                                                                                                                                      
  \end{tabular}
  \label{tab:exp-dataset-overview}
  \vspace{-1.5 em}
  
\end{table}

\section{Method}
\label{sec:method}

In this work, we aim to develop a deep learning (DL)-based model
for predicting crop yields at the county level. 
Our goals are twofold. 
First, we plan to capture the meteorological effect, 
including growing season weather variations and climate change, on crop yields.
Second, we aim to leverage high-resolution satellite images for precise agricultural tracking.
The aforementioned three data sources are utilized for achieving our goals.

\subsection{Problem Statement}
\label{sec:ps}
We consider the combination of four types of data 
$(\bm{x}, \bm{y}_{s}, \bm{y}_{l}, \bm{z})$ for predicting the crop yield at each single U.S. county.
Specifically, $\bm{x} \in \mathbb{R}^{T \times G \times H \times W \times C}$ 
represents satellite images obtained from Sentinel-2 imagery, 
which capture the information of field crops on the ground.
$T$ and $G$ indicate the numbers of temporal and spatial data, respectively,
whereas $H$, $W$, and $C$ are the height, width, and number of channels in the satellite image.
$\bm{y}_{s} \in \mathbb{R}^{T \times G \times N_{1} \times d_{y}}$ 
and $\bm{y}_{l} \in \mathbb{R}^{T \times N_{2} \times d_{y}}$
are meteorological parameters
obtained from the HRRR dataset,
representing the short-term and the long-term historical data, respectively.
Here, $N_{1}$ and $N_{2}$ are the numbers of daily and monthly HRRR data points, respectively,
and $d_{y}$ indicates the number of weather parameters.
Note that the short-term meteorological data 
is the daily HRRR data during the growing season, 
while the long-term historical meteorological data is the monthly HRRR data from the past several years 
(\eg, $2018$ to $2020$ for predicting crop yields in $2021$).
$\bm{z} \in \mathbb{R}^{d_{z}}$ is the ground-truth crop information
obtained from the USDA dataset,
with $d_{z}$ representing the number of parameters for the crop data.

\begin{figure*}[!t]
    \centering
    \includegraphics[width=.75
    \textwidth]{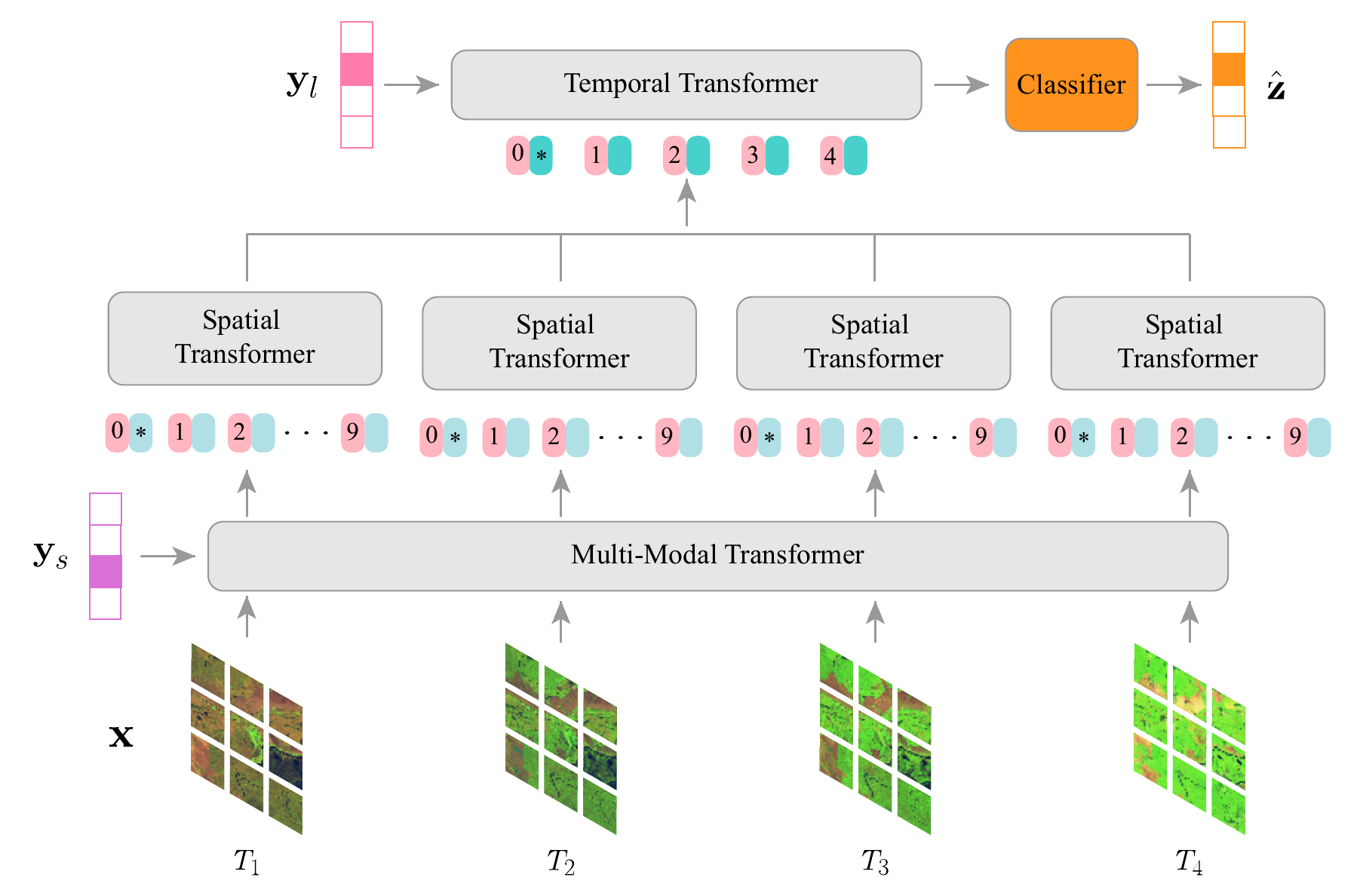}
      \vspace{-0.5 em}
    \caption{
        The architecture of our proposed MMST-ViT.
    }
    \label{fig:arch}
    \vspace{-1.5 em}
\end{figure*}

\subsection{Challenges}
\label{sec:ps}
Three challenges exist upon achieving our goals, outlined as follows.

\par\smallskip\noindent
{\bf Capturing the Effect of Growing Season Weather Variations on Crop Growth.}
Previous studies only consider the meteorological data (or the remote sensing data)
for crop yield predictions, overlooking the direct impact of growing season weather variations on crop growth.
To capture such an impact, we aim to leverage both visual remote sensing and numerical meteorological data.
In particular, the Multi-Head Attention~(MHA) technique~\cite{vaswani:nips17:attention} 
is adopted to capture the meteorological effects on crop growth.
However, how to perform multi-modal attention over visual remote sensing data
and numerical meteorological data remains open.

\par\smallskip\noindent
{\bf Lacking a Mechanism for Pre-training Multi-Modal Model.} 
Deep neural networks~(DNNs), 
especially Vision Transformer (ViT)-based models, 
are prone to overfitting,
requiring appropriate pre-training to achieve satisfactory performance.
Unfortunately, conventional pre-training techniques (\eg, SimCLR~\cite{chen:icml20:simclr}) 
only marginally improve crop yield prediction performance
as they consider the visual data only, ineffective for pre-training multi-modal models.
How to pre-train the multi-modal model for satisfactory crop yield predictions 
is challenging to be addressed.

\par\smallskip\noindent
{\bf Capturing the Impact of Climate Change on Crops.} 
As reported by a prior study~\cite{rolnick:survey23:climate_change},
the long-term climate change in the atmosphere would gradually decrease the crop yield on the ground.
Some studies~\cite{khaki2020cnn,fan:aaai23:crop_prediction} have demonstrated that 
taking the climate change effect into account can better crop yield prediction.
But their designs overlook the remote sensing data,
viewed as essential for precise agricultural tracking.
So far, how to develop effective ways to capture the impact of long-term climate change 
on crop yields is challenging and unanswered yet.

\subsection{Our Proposed Approach}
\label{sec:ps}

To tackle the aforementioned challenges,
we develop the \underline{M}ulti-\underline{M}odal \underline{S}patial-\underline{T}emporal \underline{Vi}sion \underline{T}ransformer~(MMST-ViT)
for predicting crop yields at the county level,
by leveraging the remote sensing data and the short-term and long-term meteorological data.

\par\smallskip\noindent
{\bf Model Overview.} 
Our proposed MMST-ViT consists of three key components,
\textit{i.e.}, Multi-Modal Transformer, Spatial Transformer, and Temporal Transformer,
as shown in Figure~\ref{fig:arch}.
The Multi-Modal Transformer is designed to capture the impact of short-term meteorological variations on crop growth,
by leveraging the satellite images (\textit{i.e.}, $\bm{x}$) 
and the short-term meteorological parameters (\textit{i.e.}, $\bm{y}_{s}$).
The Spatial Transformer then utilizes the output of the Multi-Modal Transformer
for learning the global spatial information of a county.
Next, the Temporal Transformer combines the outputs of our Spatial Transformer 
and the long-term meteorological data (\ie, $\bm{y}_{l}$) 
to capture both global temporal information and the impact of long-term climate change on crop yields. 
Finally, the output of our Temporal Transformer
is used by a linear classifier for predicting the annual crop yields.
The details of each component are provided below.

\par\smallskip\noindent
{\bf Multi-Modal Transformer.} 
Aiming to capture the direct impact of 
atmospheric weather variations on crop growth, 
the Multi-Modal Transformer consists of a visual backbone network and a multi-modal attention layer.
The former extracts high-quality visual representations 
from satellite images for accurate agricultural tracking,
while the latter captures the relationship between the visual representation and meteorological parameters, 
to understand the impact of meteorological parameters on crop growth.
Let $f_{\bm{\theta}}$: ($ \mathbb{R}^{T \times G \times H \times W \times C}, 
~\mathbb{R}^{T \times G \times N_{1} \times d_{y}}) \rightarrow  \mathbb{R}^{T \times G \times d}$ 
be our Multi-Modal Transformer, 
with $\bm{\theta}$ denoting the parameters for DNNs
and $d$ representing the dimension for hidden vectors.
Notably, all the hidden dimensions in this paper are set to the same size of $d$.
As such, the proposed Multi-Modal Transformer can be expressed as
$f_{\bm{\theta}} (\bm{x}, \bm{y}_{s}) = \bm{v}_{m}$,
where $ \bm{x} $ and $ \bm{y}_{s}$
are the Sentinel-2 images and the short-term meteorological data, respectively.
$ \bm{v}_{m} \in \mathbb{R}^{T \times G \times d}$ is the output of our Multi-Modal Transformer.

In this work, we utilize Pyramid Vision Transformer~(PVT)~\cite{wang:iccv21:pvt} as the visual backbone network
since it advances ResNet~\cite{he:cvpr16:resnet} and vanilla ViT~\cite{dosovitskiy:iclr21:vit} 
respectively by having the global receptive field and by enabling high-resolution feature maps.
To capture the direct impact of meteorological parameters on crop growth during the growing season,
a naive way is by concatenating visual and numerical representations 
extracted respectively from the Sentinel-2 images and the meteorological parameters.
Unfortunately, our empirical results show that such a naive way cannot achieve satisfactory performance.
Inspired by the recent success of multi-head attention mechanisms
\cite{vaswani:nips17:attention,dosovitskiy:iclr21:vit,rombach:cvpr22:stable_diffision},
we devise a novel Multi-Modal Multi-Head Attention (MM-MHA) mechanism
to capture the impact of meteorological parameters on crop growth, 
mathematically expressed as 
\begin{equation} \label{eq:mm-attn}
    \begin{gathered}
        \textrm{MM-MHA} (\bm{Q}, \bm{K}, \bm{V}) = \textrm{Softmax} (\bm{Q} \bm{K}^{T} / \sqrt{d}) \bm{V},  \\
        \bm{Q} = \bm{W}^{Q}_{m} \cdot \varphi_{m} (\bm{x}), ~\bm{K} = \bm{W}^{K}_{m} \cdot \pi_{m} (\bm{y}_{s}), \\ 
        \bm{V} = \bm{W}^{V}_{m} \cdot \pi_{m} (\bm{y}_{s}).       
    \end{gathered}
\end{equation}
Here, $\varphi_{m} (\bm{x}) \in \mathbb{R}^{T \times G \times N_{p} \times d}$ is the visual representation encoded by PVT, 
with $N_{p}$ indicating the number of image patches.
$\pi_{m} (\bm{y}_{s}) \in \mathbb{R}^{T \times G \times N_{1} \times d}$ is the meteorological parameters after the linear projection.
Notably, $\bm{W}^{Q}_{m}, \bm{W}^{K}_{m},$ and  $\bm{W}^{V}_{m}$ are learnable projection matrices,
similar to those in prior studies~\cite{vaswani:nips17:attention,rombach:cvpr22:stable_diffision}.
To our knowledge, this is the first attempt at developing a multi-modal multi-attention approach 
for leveraging both visual remote sensing data and numerical meteorological data.

\begin{figure}[!t]
    \centering
    \includegraphics[width=.48\textwidth]{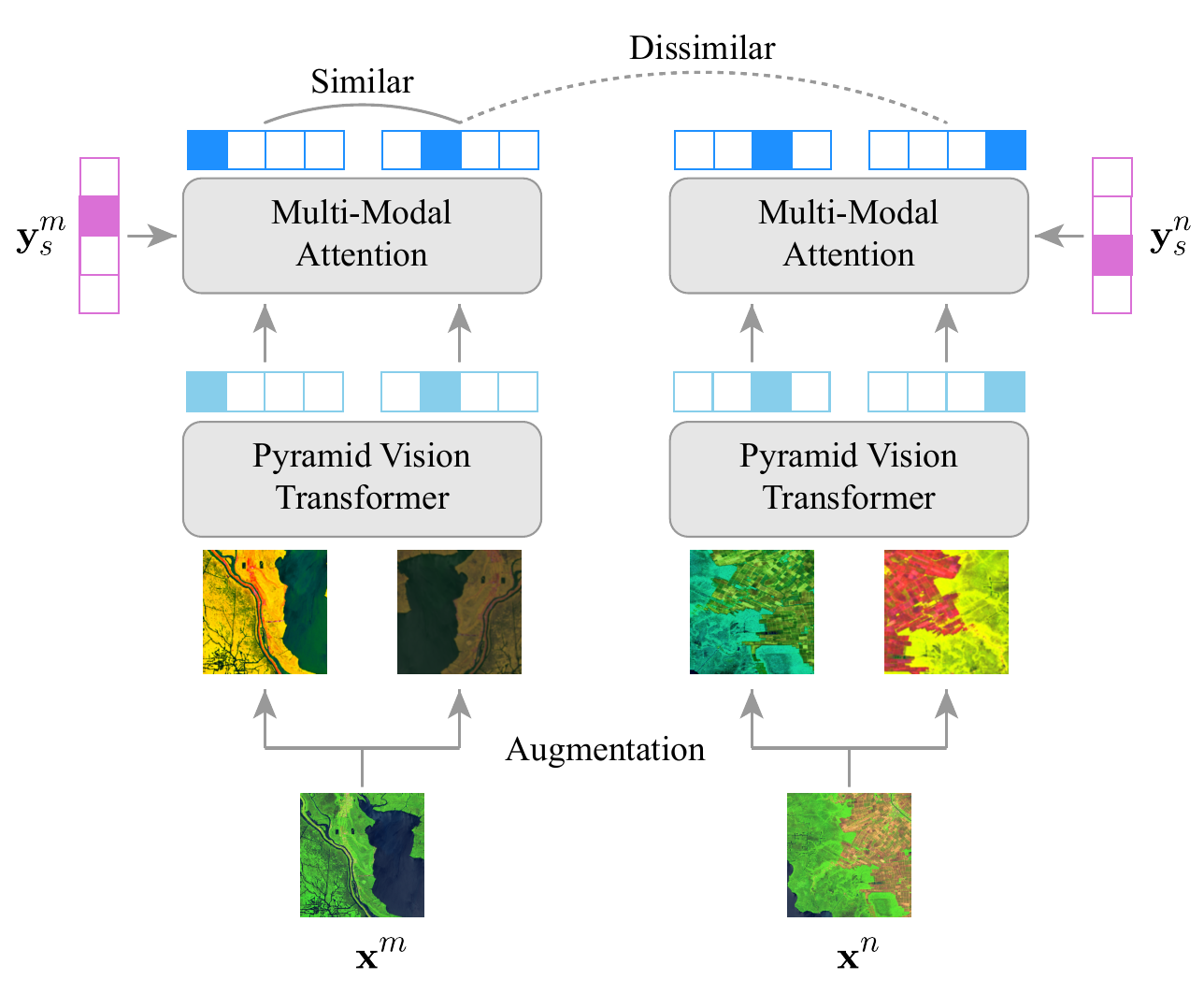}
    \caption{
        Multi-modal self-supervised learning.
    }
    \label{fig:method-pretrain}
    \vspace{-1.0 em}
\end{figure}

However, ViT-based models are highly sensitive to overfitting,
calling for pre-training with millions of visual data~\cite{dosovitskiy:iclr21:vit}.
Meanwhile, real-world crop yield prediction usually lacks sufficient crop data for pre-training. 
Worse still, capturing the impact of short-term weather conditions on  crops 
requires considering both the visual and the numerical data simultaneously.
Although self-supervised learning techniques~\cite{chen:icml20:simclr,bao:iclr22:beit,he:cvpr22:mae}
have recently been developed for pre-training DNNs, 
they cannot tackle the aforementioned issue at the same time.
For example, SimCLR~\cite{chen:icml20:simclr} fails to address the multi-modal data issue.

Here, we propose a novel multi-modal self-supervised learning 
for pre-training our Multi-Modal Transformer without human supervision, which is inspired by SimCLR, but having two differences.
First, we utilize the PVT instead of convolutional neural networks~(CNNs) as the backbone network
since PVT advances CNNs by having a global receptive field~\cite{wang:iccv21:pvt}.
Second, we replace the Multi-Layer Perceptron (MLP) layer in SimCLR 
with our proposed multi-modal self-attention layer~(\textit{i.e.}, ~\eqref{eq:mm-attn})
for simultaneously tackling visual and numerical data.
Figure~\ref{fig:method-pretrain} illustrates its architecture.
Given a satellite image $\bm{x}^{m}$ (or $\bm{x}^{n}$),
we perform the data augmentation to get its two augmented images,
and then feed them to the PVT to get two sets of visual representations.
After that, we use our proposed MM-MHA~(\ie, \eqref{eq:mm-attn})
to perform multi-modal attention between the visual representations 
and the corresponding short-term parameters 
$\bm{y}^{m}_{s}$ (or $\bm{y}^{n}_{s}$),
arriving at two sets of output sequences.
Here, we regard two sets of output sequences out of a given satellite image (\eg, $\bm{x}^{m}$) as the positive pair.
%
Meanwhile,  two sets of output sequences from different satellite images
(\eg, $\bm{x}^{m}$ and $\bm{x}^{n}$) are regarded as the negative pair.
Similar to vanilla ViT~\cite{dosovitskiy:iclr21:vit},
the head of output sequences (\ie, the classification token)
is taken as the output of our Multi-Modal Transformer,
\ie, the hidden vector $\bm{v}_{m}$.
As such, our multi-modal contrastive loss is defined as,
\begin{equation} \label{eq:loss-sim-clr}
    \begin{aligned}
      \ell (i,j) &= - \log \frac{\exp(\text{sim}(\bm{v}_{m}^{i}, \bm{v}_{m}^{j}) / \tau )}
      {\sum_{i \neq k, k = 1, \dots, 2B} \exp (\text{sim}(\bm{v}_{m}^{i}, \bm{v}_{m}^{k}) / \tau  ) }, \\
      & \mathcal{L}_{\rm{cl}} = \frac{1}{2B} \sum_{k=1}^{B} \left[ \ell(k, k+B) +\ell(k+B, k) \right],  
    \end{aligned}  
\end{equation}
where $B = T \times G$.
Here, $\bm{v}_{m}^{i}$ and $\bm{v}_{m}^{j}$ are the positive pair,
and $\tau$ is the temperature parameter.
Intuitively, our \eqref{eq:loss-sim-clr} encourages two hidden vectors from the same satellite image
(\ie, representing the same region) to be similar,
and two hidden vectors from different satellite images (\ie, representing different regions) 
to be dissimilar.
As such, we can pre-train our Multi-Modal Transformer to 
capture the impact of short-term meteorological parameters on crop growth 
without labor-intensive human supervision.

\par\smallskip\noindent
{\bf Spatial Transformer.}
Recall that the Sentinel-2 Imagery dataset of a county is partitioned into 
multiple fine-grained grids for precise agricultural tracking.
To learn spatial dependency among those grids,
we propose a Spatial Transformer
$g_{\phi}: \mathbb{R}^{T \times G \times d} \rightarrow \mathbb{R}^{T \times d}$,
whose purpose is to capture the global spatial representations for counties.
The design of our Spatial Transformer is inspired by the vanilla ViT~\cite{dosovitskiy:iclr21:vit}
but with a flexible number of positional embeddings.
It is used to tackle the scenario that the number of grids varies among counties
as it depends on county sizes.
To learn global spatial information for counties,
the Spatial Transformer prepends a learnable classification token 
$\bm{v}_{m}^{\textrm{cls}} \in \mathbb{R}^{T \times 1 \times d} $, arriving at 
\begin{equation} \label{eq:s-embed}
    \begin{aligned}
        \varphi_{s} (\bm{v}_{m}) = [\bm{v}_{m}^{\textrm{cls}};~\bm{v}_{m}^{1};~\bm{v}_{m}^{2}; ~\dots~ ; ~\bm{v}_{m}^{G}] + \mathbf{E}_{\textrm{pos}}, 
    \end{aligned}
\end{equation}
where $\mathbf{E}_{\textrm{pos}} \in  \mathbb{R}^{T \times (G+1) \times d}$ is the flexible positional embeddings.
We propose a Spatial Multi-Head Attention (S-MHA)
to learn spatial dependency among grids.
Mathematically,
\begin{equation} \label{eq:sp-attn}
    \begin{gathered}
        \textrm{S-MHA} (\bm{Q}, \bm{K}, \bm{V}) = \textrm{Softmax} (\bm{Q} \bm{K}^{T} / \sqrt{d}) \bm{V}, \\
        \bm{Q} = \bm{W}^{Q}_{s}  \cdot \varphi_{s} (\bm{v}_{m}), ~\bm{K} = \bm{W}^{K}_{s} \cdot \varphi_{s} (\bm{v}_{m}),  \\
        \quad \bm{V} = \bm{W}^{V}_{s} \cdot \varphi_{s} (\bm{v}_{m}).  
    \end{gathered}
\end{equation}
Here, $\bm{W}^{Q}_{s}$, $\bm{W}^{K}_{s}$, $\bm{W}^{V}_{s}$ are three learnable matrices, 
similar to those given in \eqref{eq:mm-attn}.
Finally, we regard the head of output sequences as global spatial representations of counties.
%
That is, we have $g_{\bm{\phi}} (\bm{v}_{m}) = \bm{v}_{s}$,
with $\bm{v}_{s} \in \mathbb{R}^{T \times d}$ signifying the hidden vector
that incorporates global spatial information.

\par\smallskip\noindent
{\bf Temporal Transformer.} 
The Temporal Transformer possesses two goals.
The first goal aims to learn temporal dependency among the hidden vectors $\bm{v}_{s}$,
\ie, the outputs of our Spatial Transformer.
The second goal is to capture the effect of long-term climate change on crop yields.
To achieve the first goal,
we add a classification token $\bm{v}^{\textrm{cls}}_{s}$ for 
learning the global temporal representation, \ie,
\begin{equation} \label{eq:t-embed}
    \begin{aligned}
        \varphi_{t} (\bm{v}_{s}) = [\bm{v}_{s}^{\textrm{cls}};~\bm{v}_{s}^{1};~\bm{v}_{s}^{2}; ~\dots~ ; ~\bm{v}_{s}^{T}] + \mathbf{E}_{\textrm{tmp}}, 
    \end{aligned}
\end{equation}
where $\mathbf{E}_{\textrm{tmp}} \in  \mathbb{R}^{(T+1) \times d}$ is the temporal embedding,
similar to the positional embeddings $\mathbf{E}_{\textrm{pos}}$ expressed in \eqref{eq:s-embed}. 
The novel Temporal Multi-Head Attention (T-MHA) 
is also devised to capture the impact of long-term historical meteorological parameters on crop yields.
Its key idea is to incorporate a relative meteorological bias into each head for similarity computation,
motivated by prior studies~\cite{bao:icml20:UniLMv2,raffel:jmlr20:exploring,liu:iccv21:swin}.
Mathematically, our T-MHA can be expressed by
\begin{equation} \label{eq:t-attn}
    \begin{gathered}
        \textrm{T-MHA} (\bm{Q}, \bm{K}, \bm{V}) = \textrm{Softmax} (\bm{Q} \bm{K}^{\bm{T}} / \sqrt{d} + \pi_{t} (\bm{y}_{l}) ) \bm{V}, \\
        \bm{Q} = \bm{W}^{Q}_{t} \cdot \varphi_{t} (\bm{v}_{s}), ~\bm{K} = \bm{W}^{K}_{t} \cdot \varphi_{t} (\bm{v}_{s}), \\ 
        \bm{V} = \bm{W}^{V}_{t} \cdot \varphi_{t} (\bm{v}_{s}).       
    \end{gathered}
\end{equation}
Here, $\bm{y}_{l}$ represents the long-term meteorological parameters,
and $\pi_{t} (\cdot)$ is a linear projection layer.
Similarly, $\bm{W}^{Q}_{t}$, $\bm{W}^{K}_{t}$, $\bm{W}^{V}_{t}$ are three learnable matrices.
In summary, our Temporal Transformer $h_{\bm{\psi}}: (\mathbb{R}^{T \times d}, ~\mathbb{R}^{T \times N_{2} \times d_{y}}) \rightarrow \mathbb{R}^{d}$
can be defined as $h_{\bm{\psi}} (\bm{v}_{s}, \bm{y}_{l}) = \bm{v}_{t}$, 
with $\bm{v}_{t} \in \mathbb{R}^{d}$ being the hidden vector 
that incorporates both global temporal information
and the impact of climate change on crops simultaneously.

Finally, the hidden vector $\bm{v}_{t}$ is fed into a linear classifier for crop yield predictions,
\ie, $\hat{\bm{z}} = \bm{W}^{T} \bm{v}_{t} + \bm{b}$, with $\bm{W}$ and $\bm{b}$ 
respectively denoting the weights and the bias for the classifier,
and $\hat{\bm{z}} \in \mathbb{R}^{d_{z}}$ indicating the crop yield prediction.
%


\section{Experiments}
\label{sec:exp}

We conduct experiments for the county-level crop yield predictions across four U.S. states,
\ie, Mississippi (MS), Louisiana (LA), Iowa (IA), and Illinois (IL). 
Four  types of crops,
\ie, corn, cotton, soybean, and winter wheat, are taken into account for performance evaluation.

\subsection{Experimental Settings}
\label{sec:exp-setup}

\par\smallskip\noindent
{\bf Datasets.} 
We utilize the Sentinel-2 imagery and the daily HRRR data
datasets during the growing season respectively as the remote sensing data and the short-term meteorological data.
Meanwhile, the monthly HRRR data from the previous three years are used 
as the long-term meteorological parameters.

\par\smallskip\noindent
{\bf Compared Approaches.} 
We compare  our proposed MMST-ViT to three DL-based approaches,
with one, \ie,  \textbf{ConvLSTM}~\cite{shi:nips15:conv_lstm}, developed for spatial-temporal prediction, 
and the other two, \ie, \textbf{CNN-RNN}~\cite{khaki2020cnn} and \textbf{GNN-RNN}~\cite{fan:aaai23:crop_prediction},
developed for crop yield predictions.
Minor rectifications are made to them so that they can admit the Sentinel-2 imagery and HRRR datasets as their inputs.
Other hyperparameters, unless specified otherwise, are set to the values as those reported in their original studies.

\par\smallskip\noindent
{\bf Metrics.} 
We take three performance metrics, \textit{i.e.}, \textbf{Root Mean Square Error (RMSE)}, 
\textbf{R-squared (R$^{2})$}, 
and \textbf{Pearson Correlation Coefficient (Corr)},
for comparative evaluation. 
Notably, a lower value of RMSE or a higher value of R$^{2}$ (or Corr) indicates better performance.
%

%
\begin{table*} [!t]
	\scriptsize
    \centering
    \setlength\tabcolsep{8 pt}
    \caption{Model details used in our study}
    \vspace{-0.5em}
    \begin{tabular}{@{}|cc|c|c|c|c|c|@{}}
        \toprule
        \multicolumn{2}{|c|}{Model}                         & Layer & Hidden Size & Head & MLP Size & Others            \\ \midrule
        \multicolumn{1}{|c|}{\multirow{2}{*}{ \text{Multi-Modal Transformer}}} &
          PVT-T/4 &
          \{2, 2, 2, 2\} &
          \{64, 128, 320, 512\} &
          \{1, 2, 5, 8\} &
          \{512, 1024, 1280, 2048\} &
         SR Ratios = \{8, 4, 2, 1\} \\ 
        \multicolumn{1}{|c|}{}                     & MM-MHA & 2     & 512         & 8    & 2048     & Context Size = 9 \\ 
        \multicolumn{1}{|c|}{\text{Spatial Transformer}}  & S-MHA  & 4     & 512         & 3    & 2048     & -                 \\ 
        \multicolumn{1}{|c|}{\text{Temporal Transformer}} & T-MHA  & 4     & 512         & 3    & 2048     & Context Size = 9 \\ \bottomrule
    \end{tabular}
    
	\label{tab:exp-model-size}
	\vspace{-0.5 em}
	
\end{table*}

\begin{table*}[!t]
    \scriptsize
    \centering
    \setlength\tabcolsep{6 pt}
    \caption{
        Overall comparative crop yield predictions for $2021$, with the best results shown in bold.
        Cotton yields are measured in pounds per acre (LB/AC), 
        whereas other crop yields are measured in bushels per acre (BU/AC)   
        }
        \vspace{-1em}
        \begin{tabular}{@{}|c|ccc|ccc|ccc|ccc|@{}}
            \toprule
            \multirow{2}{*}{Method} & \multicolumn{3}{c|}{Corn} & \multicolumn{3}{c|}{Cotton} & \multicolumn{3}{c|}{Soybean} & \multicolumn{3}{c|}{Winter Wheat} \\ \cmidrule(l){2-13} 
             &
              \multicolumn{1}{c}{RMSE ($\downarrow$)} &
              \multicolumn{1}{c}{$\text{R}^{2}$ ($\uparrow$)} &
              Corr ($\uparrow$) &
              \multicolumn{1}{c}{RMSE ($\downarrow$)} &
              \multicolumn{1}{c}{$\text{R}^{2}$ ($\uparrow$)} &
              Corr ($\uparrow$) &
              \multicolumn{1}{c}{RMSE ($\downarrow$)} &
              \multicolumn{1}{c}{$\text{R}^{2}$ ($\uparrow$)} &
              Corr ($\uparrow$) &
              \multicolumn{1}{c}{RMSE ($\downarrow$)} &
              \multicolumn{1}{c}{$\text{R}^{2}$ ($\uparrow$)} &
              Corr ($\uparrow$) \\ \midrule
            ConvLSTM &
              \multicolumn{1}{c}{18.6} &
              \multicolumn{1}{c}{0.611} &
              0.782 &
              \multicolumn{1}{c}{65.4} &
              \multicolumn{1}{c}{0.715} &
              0.846 &
              \multicolumn{1}{c}{7.2} &
              \multicolumn{1}{c}{0.616} &
              0.785 &
              \multicolumn{1}{c}{7.4} &
              \multicolumn{1}{c}{0.511} &
              0.715 \\ 
            CNN-RNN &
              \multicolumn{1}{c}{14.6} &
              \multicolumn{1}{c}{0.705} &
              0.839 &
              \multicolumn{1}{c}{69.5} &
              \multicolumn{1}{c}{0.653} &
              0.808 &
              \multicolumn{1}{c}{5.8} &
              \multicolumn{1}{c}{0.703} &
              0.839 &
              \multicolumn{1}{c}{7.5} &
              \multicolumn{1}{c}{0.614} &
              0.783 \\ 
            GNN-RNN &
              \multicolumn{1}{c}{14.2} &
              \multicolumn{1}{c}{0.730} &
              0.854 &
              \multicolumn{1}{c}{58.5} &
              \multicolumn{1}{c}{0.647} &
              0.804 &
              \multicolumn{1}{c}{5.4} &
              \multicolumn{1}{c}{0.748} &
              0.865 &
              \multicolumn{1}{c}{6.0} &
              \multicolumn{1}{c}{0.621} &
              0.788 \\ \midrule
              \textbf{Ours} &
              \multicolumn{1}{c}{\textbf{10.5}} &
              \multicolumn{1}{c}{\textbf{0.811}} &
              \textbf{0.900} &
              \multicolumn{1}{c}{\textbf{42.4}} &
              \multicolumn{1}{c}{\textbf{0.790}} &
              \textbf{0.889} &
              \multicolumn{1}{c}{\textbf{3.9}} &
              \multicolumn{1}{c}{\textbf{0.843}} &
              \textbf{0.918} &
              \multicolumn{1}{c}{\textbf{4.6}} &
              \multicolumn{1}{c}{\textbf{0.785}} &
              \textbf{0.886} \\ \bottomrule
            \end{tabular}
    
    \label{tab:exp-crop-pred}
    \vspace{-1em}
    
\end{table*}

\par\smallskip\noindent
{\bf Model Size.} 
Our proposed MMST-ViT builds on the top of Vision Transformer (ViT)~\cite{dosovitskiy:iclr21:vit}.
Similar to ViT, it consists of a stack of Transformer blocks~\cite{vaswani:nips17:attention},
where each Transformer block includes a Multi-Head Attention (MHA) block and an MLP block, 
and both of which incorporate the LayerNorm~\cite{ba:arxiv16:ln} for normalization.
Table~\ref{tab:exp-model-size} presents 
the details of model sizes used in our experiments. 
Note that ``PVT-T/4'' refers to the PVT-Tiny model with a patch size of 4, 
and ``SR Ratios'' represents the spatial reduction ratio in PVT. 
Following the original PVT~\cite{wang:iccv21:pvt}, 
we divide the PVT backbone into four stages 
and report the corresponding model sizes at each stage in the second row of Table~\ref{tab:exp-model-size}. 
The other transformers are single-stage, similar to the vanilla ViT~\cite{dosovitskiy:iclr21:vit}. 
Notably, ``Context Size'' represents the number of meteorological parameters used in our study.

\par\smallskip\noindent
{\bf Hyperparameters.} 
Our model is pre-trained for $200$ epochs using AdamW~\cite{loshchilov:iclr19:adamw} 
with $\beta_{1} = 0.9$, $\beta_{2} = 0.95$, a weight decay of $0.05$, 
and a cosine decay schedule~\cite{loshchilov:iclr17:sgdr} 
with an initial learning rate of $1e-4$, and warmup epochs of $20$. 
To perform data augmentation for pre-training, 
various techniques such as random cropping, random horizontal flipping, random Gaussian blur, and color jittering are employed.
These techniques are similar to those used in SimCLR~\cite{chen:icml20:simclr}.
After pre-training, we fine-tune our proposed MMST-ViT for $100$ epochs 
using AdamW with $\beta_{1} = 0.9$, $\beta_{2} = 0.999$, 
a cosine decay schedule with an initial learning rate of $1e-3$, 
and warmup epochs of $5$.

\subsection{Comparative Performance Evaluation} 
\label{sec:exp-overall-perform}
We conduct  experiments for predicting $2021$ crop yields at the county level 
across four aforementioned U.S. states, 
with prediction performance measured by the three metrics of RMSE, R$^{2}$, and Corr.

Table~\ref{tab:exp-crop-pred} lists the comparative performance results of our MMST-ViT 
and its three counterparts, \ie, ConvLSTM, CNN-RNN, and GNN-RNN. 
Four observations are obtained from the table.
First, our approach achieves the best performance under all metrics.
In particular, for the soybean yield prediction,
our approach achieves the lowest RMSE of $3.9$ and 
the highest R$^{2}$ (or Corr) of $0.843$ (or $0.918$).
With the lowest RMSE value, 
our predicted soybean yields are the closest to the actual amounts.
In addition, the highest R$^{2}$ and Corr values demonstrate 
that our predicted soybean yields are best correlated to actual figures.
Second, our approach significantly outperforms ConvLSTM, 
with its RMSE values always lower than those of ConvLSTM markedly, 
ranging from $2.8$ (for winter wheat) to $23.0$ (for cotton).
The reason is that ConvLSTM overlooks the impact of 
meteorological parameters on crop growth.
Third, 
CNN-RNN underperforms our approach by $4.1$ (for corn), by $27.1$ (for cotton), by $1.9$ (for soybean), and by $2.9$ (for winter wheat), in terms of RMSE, 
though both models take into account the effect of long-term climate change on crop yields.
Because CNN-RNN cannot model the spatial dependency among neighborhood regions.
Fourth, MMST-ViT outperforms the most recent \sota~(\ie, GNN-RNN) measurably.  
Specifically, our RMSE value is lower by $16.1$ for cotton, 
whereas our R$^{2}$ and  Corr values are better respectively by $0.165$ and $0.098$ for winter wheat.
These results are contributed by leveraging both visual remote sensing and numerical meteorological data in our model, 
while GNN-RNN only considers the meteorological data.

\begin{figure*} [!t] 
    \captionsetup[subfigure]{justification=centering}
    \begin{subfigure}[t]{0.24\textwidth}
        \centering
        \includegraphics[width=\textwidth]{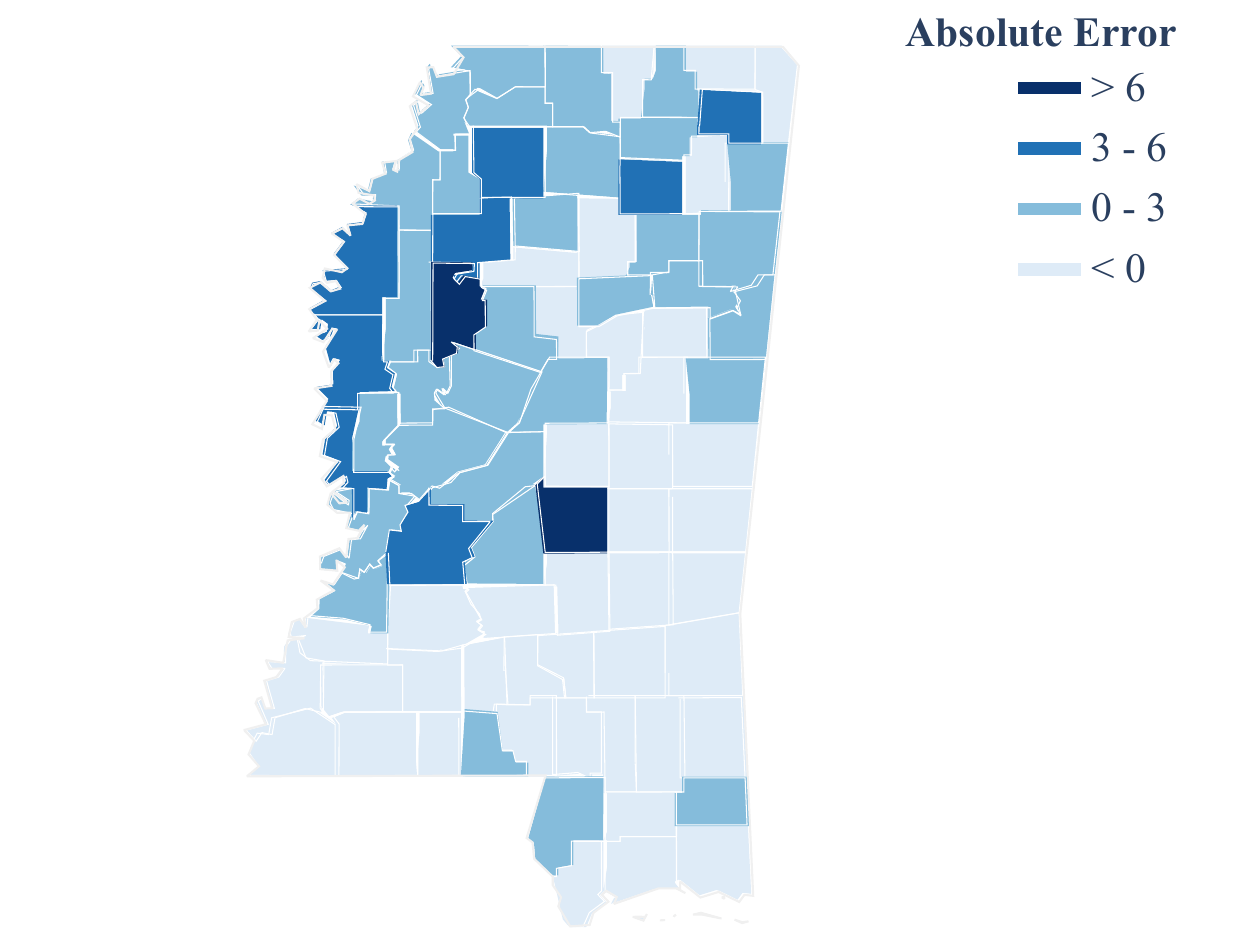}
        \caption{Mississippi (MS)}
        \label{fig:exp-viz-soybean-ms}
    \end{subfigure}
    \begin{subfigure}[t]{0.24\textwidth}
        \centering
        \includegraphics[width=\textwidth]{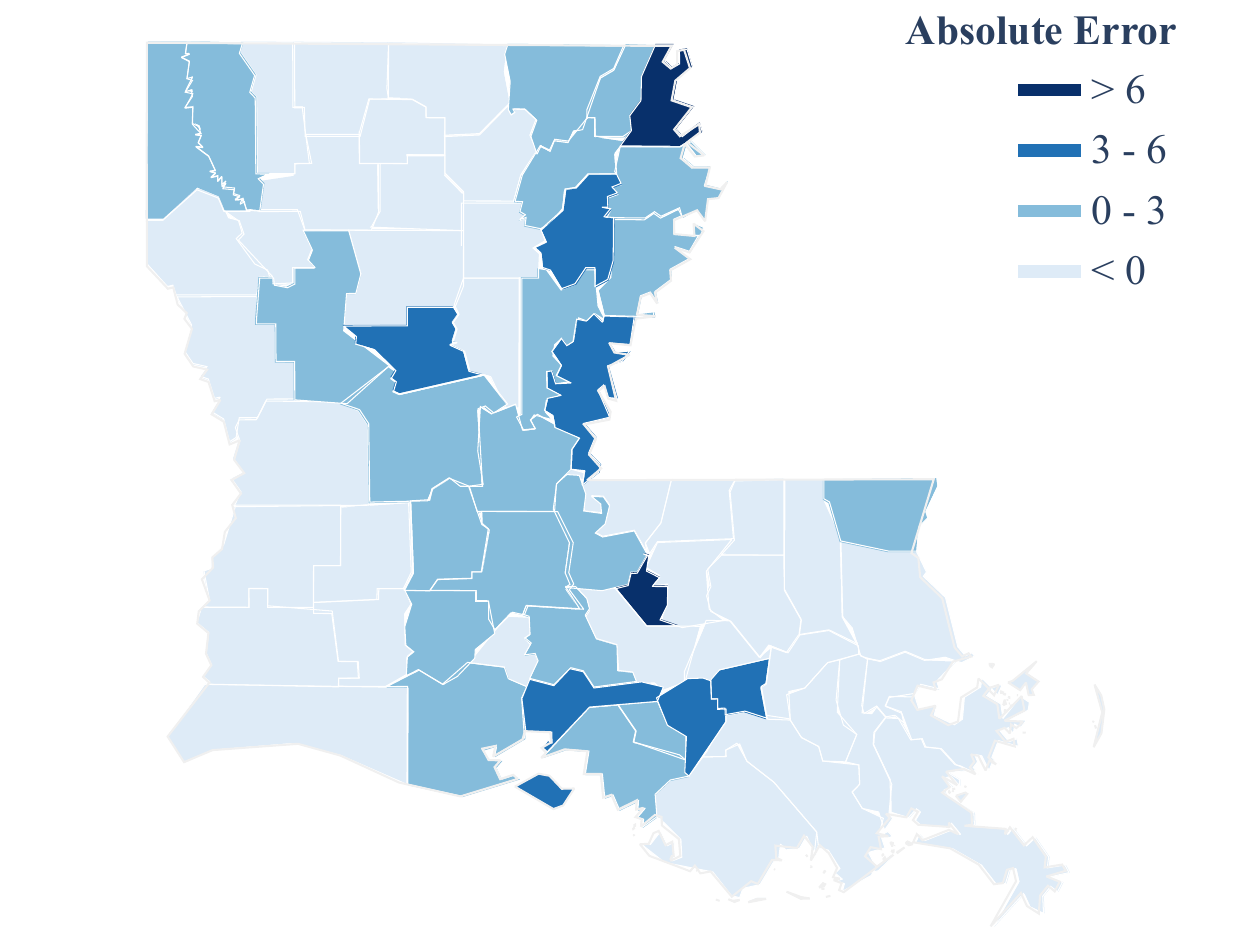}
        \caption{Louisiana (LA)}
        \label{fig:exp-viz-soybean-ls}
    \end{subfigure}
    \begin{subfigure}[t]{0.24\textwidth}
        \centering
        \includegraphics[width=\textwidth]{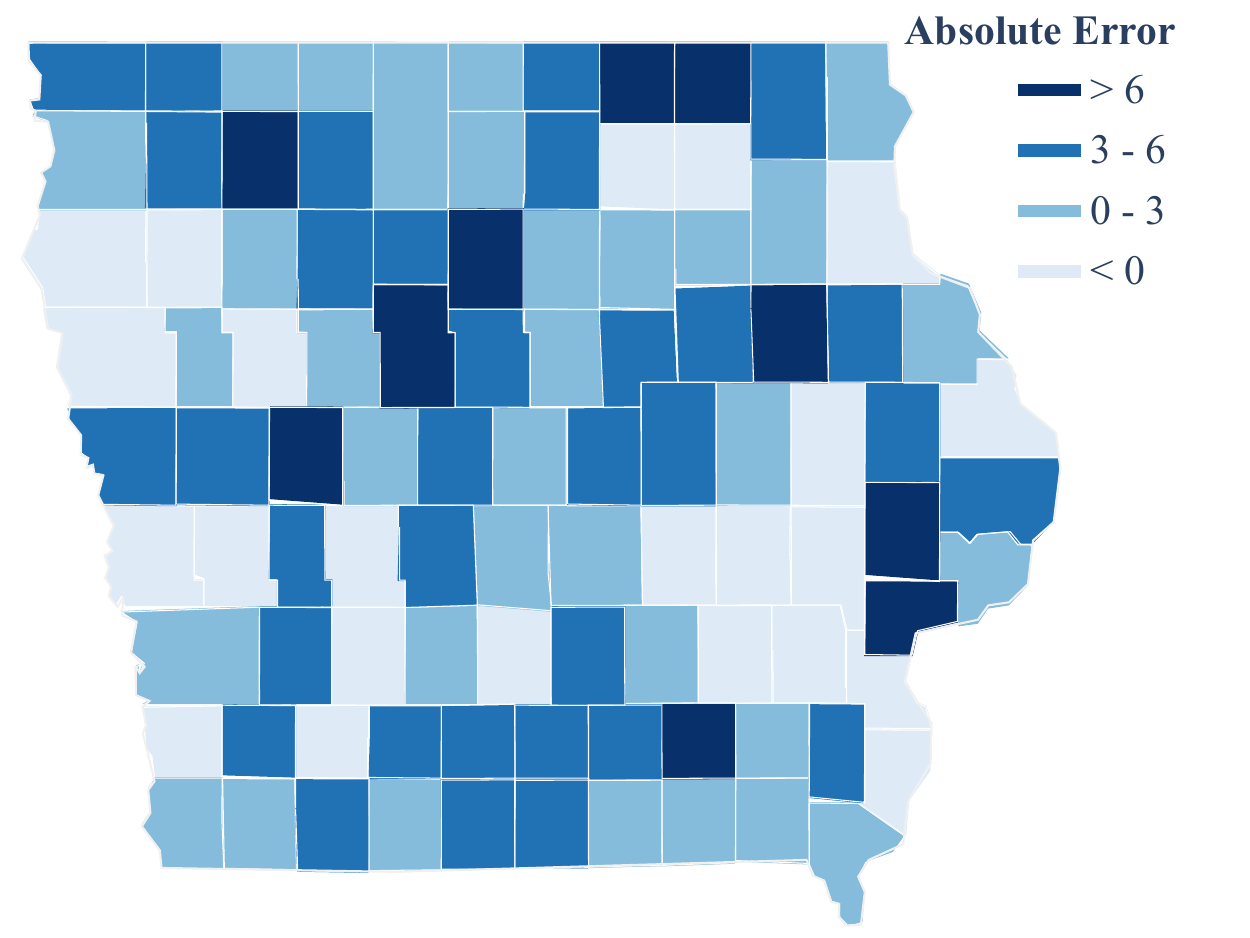}
        \caption{Iowa (IA)}
        \label{fig:exp-viz-soybean-io}
    \end{subfigure}
    \begin{subfigure}[t]{0.24\textwidth}
        \centering
        \includegraphics[width=\textwidth]{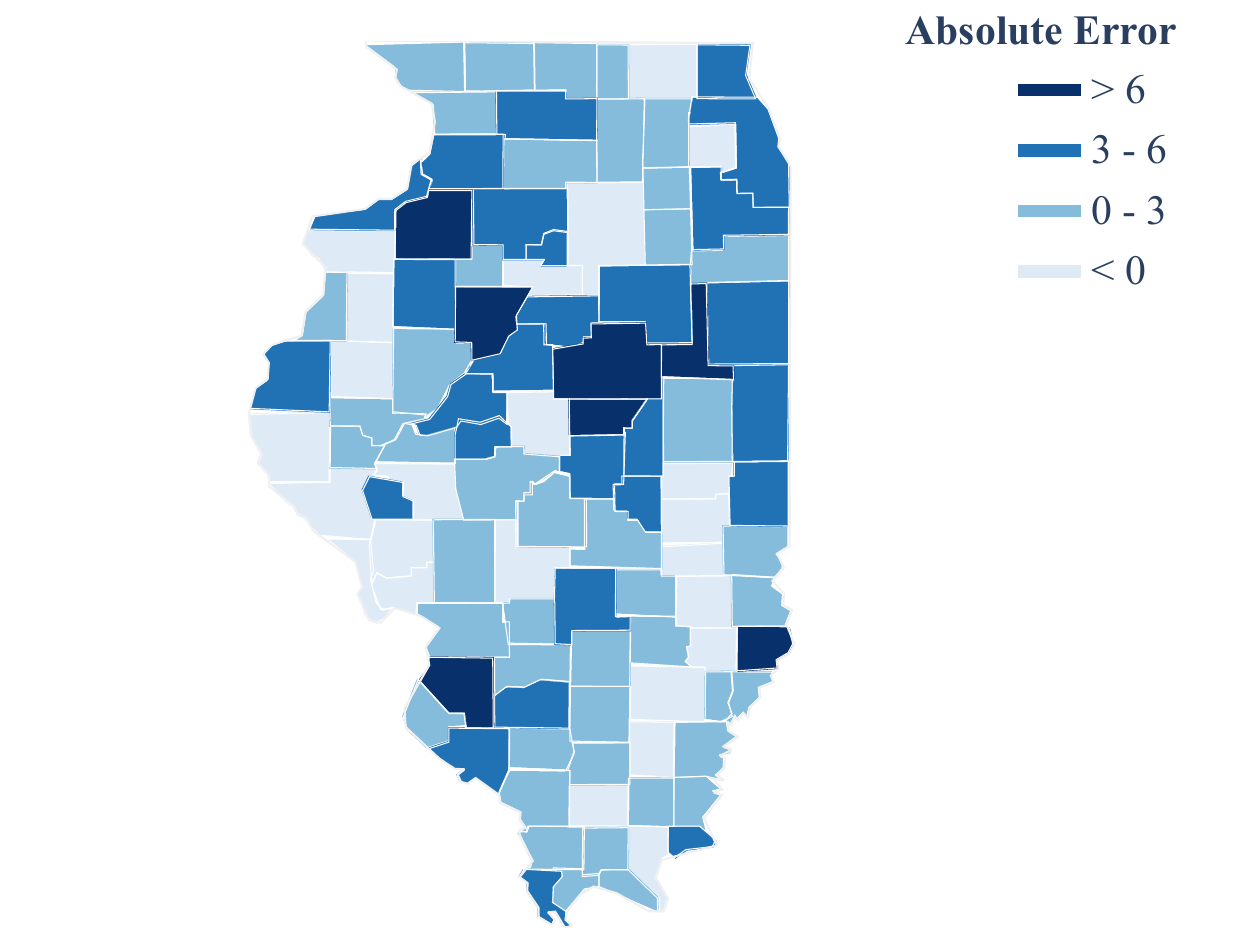}
        \caption{Illinois (IL)}
        \label{fig:exp-viz-soybean-il}
    \end{subfigure}
    \caption{
        Illustration of absolute prediction errors for soybean across four U.S. states.
        Note that a county/parish with ``$< 0$'' indicates its soybean yield data is unavailable.
    }
    \label{fig:viz-pred-error-soybean}
    \vspace{-1.0 em}
\end{figure*}

\subsection{Visualizing Crop Yield Prediction Errors}
\label{sec:exp-viz-error}
In this section, we conduct experiments to visualize crop yield prediction errors across four U.S. states: 
Mississippi (MS), Louisiana (LA), Iowa (IA), and Illinois (IL). 
We use the same experimental settings as those stated in Section~\ref{sec:exp-overall-perform}. 
Figure~\ref{fig:viz-pred-error-soybean} shows the absolute prediction errors for soybean across counties/parishes in the four states, 
with navy blue and light blue respectively indicating the high and low absolute errors. 
Our MMST-ViT model is highly effective in predicting crop yields, 
with the absolute prediction errors of $57.2 \%$ counties below $3$ BU/AC. 
Furthermore, we discover that $95.1 \%$ of counties in MS, $92.5 \%$ of parishes in LA, $86.8 \%$ of counties in IA, and $91.0 \%$ of counties in IL 
achieve decent absolute prediction errors (\ie, $\leq 6$ BU/AC). 
These empirical findings validate the robustness of MMST-ViT across various geographic locations.
%

\begin{figure*} [!t] 
    \captionsetup[subfigure]{justification=centering}
    \begin{subfigure}[t]{0.30\textwidth}
        \centering
        \includegraphics[width=\textwidth]{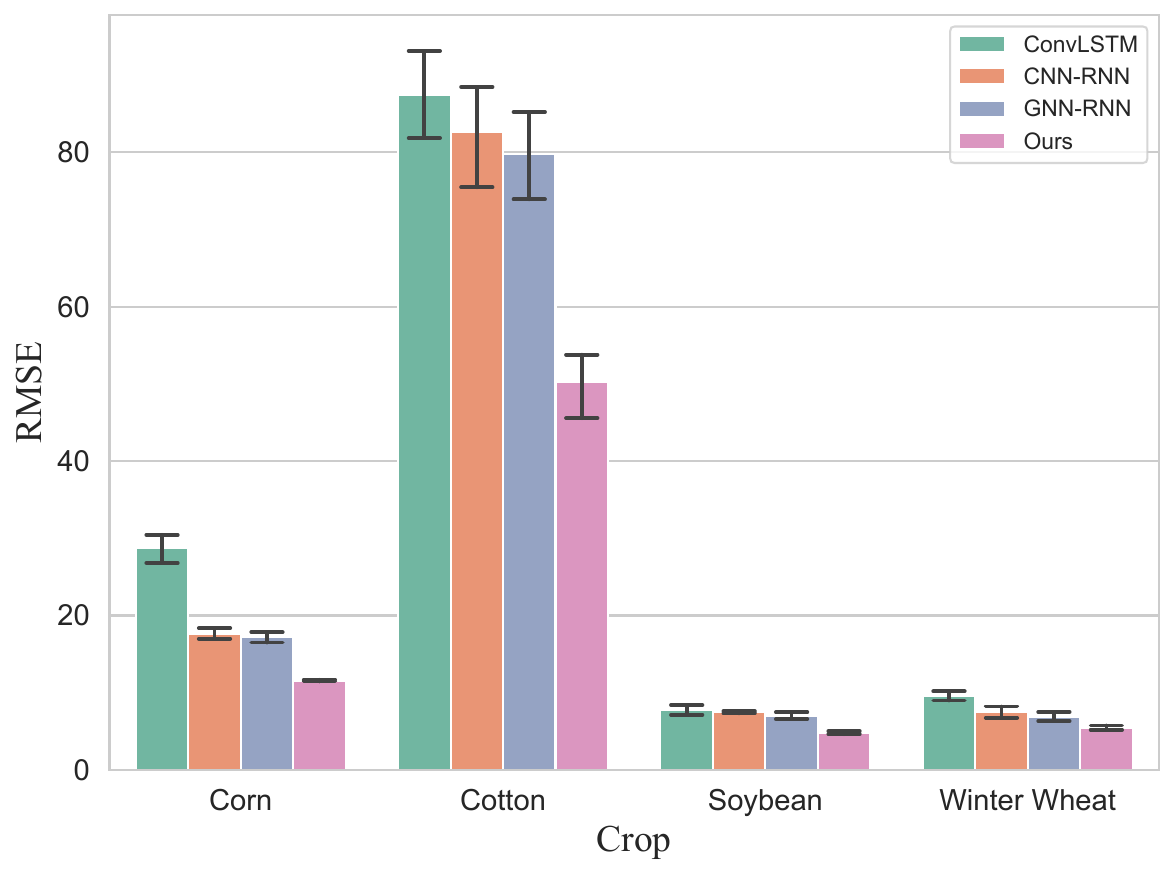}
        \caption{RMSE}
        \label{fig:exp-early-pred-rmse}
    \end{subfigure}
    \quad
    \begin{subfigure}[t]{0.30\textwidth}
        \centering
        \includegraphics[width=\textwidth]{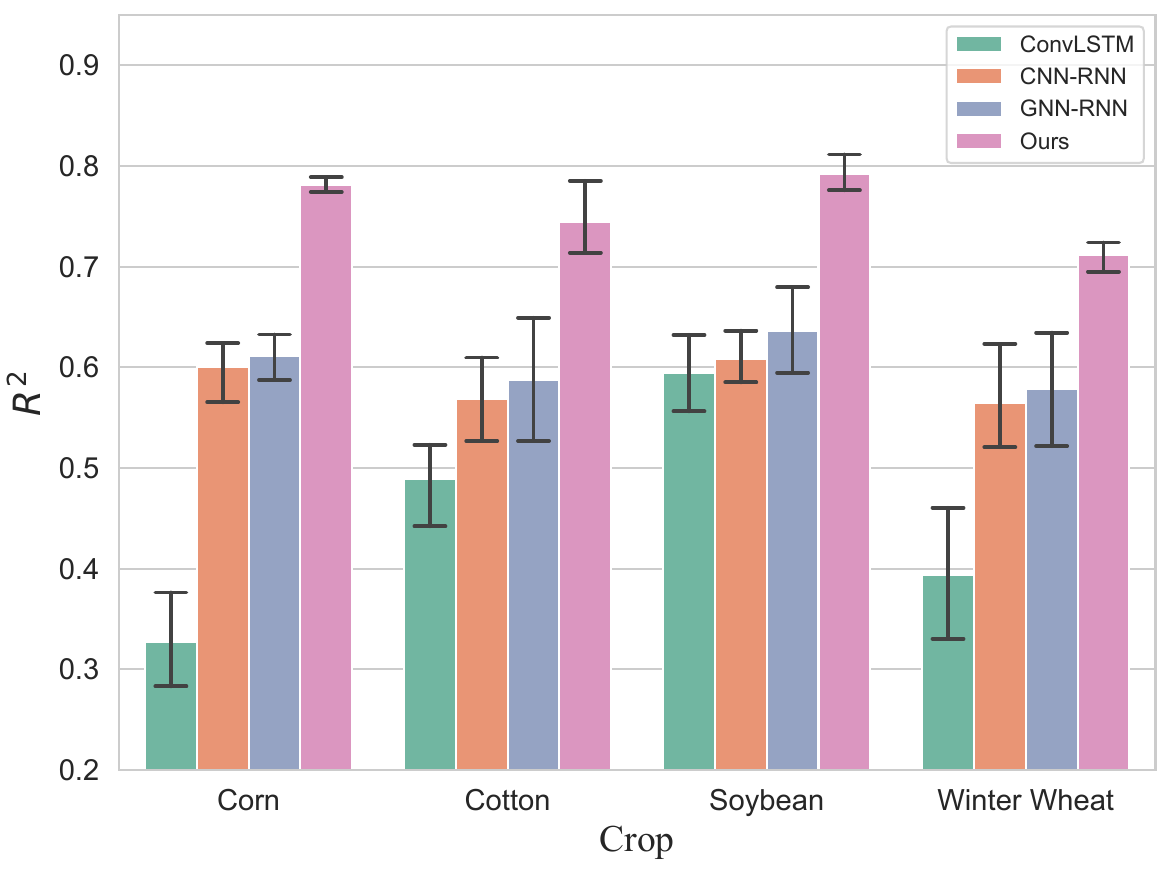}
        \caption{R$^2$}
        \label{fig:exp-early-pred-r2}
    \end{subfigure}
    \quad
    \begin{subfigure}[t]{0.30\textwidth}
        \centering
        \includegraphics[width=\textwidth]{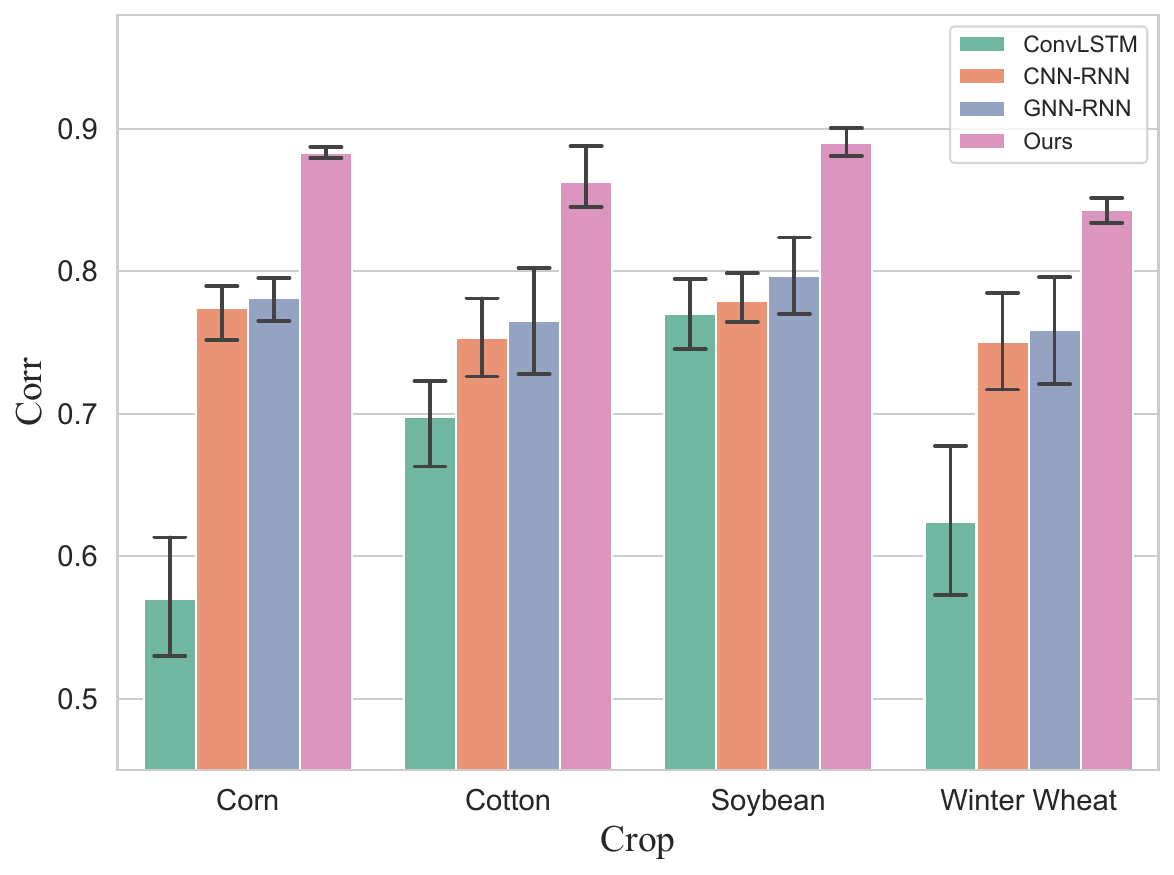}
        \caption{Corr}
        \label{fig:exp-early-pred-corr}
    \end{subfigure}
    \vspace{-0.5em}
    \caption{
    Illustration of the performance for predictions of one year ahead under (a) RMSE, (b) R$^{2}$, and (c) Corr, 
    with the cotton yield measured by LB/AC and other crop yields measured by BU/AC.
    }
    \label{fig:exp-early-pred}
    \vspace{-1.0 em}
\end{figure*}

\subsection{Performance of One Year Ahead Predictions}
\label{sec:exp-early-pred}

In practice, predicting crop yields 
well in advance of harvest, 
can provide valuable information for farmers, agribusinesses, and governments 
in their agricultural planning and decision-making processes. 
This information can be used to elevate agricultural resilience and sustainability, 
make informed financial decisions, and support global food security~\cite{horie1992yield}.
A prior study~\cite{fan:aaai23:crop_prediction} 
simulated the early prediction scenario by masking partial inputs,
successfully making predictions several months before harvest.
In this study, we attempt a step further by making crop yield predictions one year before the harvest.
In particular, we leverage remote sensing and short-term meteorological data 
during the growing season in the current year to predict crop yields in the next year.
For example, the Sentinel-2 imagery and daily HRRR data during the growing season in $2020$ 
are utilized as remote sensing data and the short-term meteorological data
for predicting crop yields in $2021$.

Figure~\ref{fig:exp-early-pred} illustrates the experimental results.
It is observed that our MMST-ViT outperforms three baseline models, 
\ie, ConvLSTM, CNN-RNN, and GNN-RNN, for the predictions of one year ahead.
Among all crops, MMST-ViT achieves the lowest RMSE values ranging from $4.7$ (for soybean) to $50.2$ (for cotton), 
the highest R$^2$ values ranging from $0.711$ (for winter wheat) to $0.792$ (for soybean), 
and the highest Corr values ranging from $0.843$ (for winter wheat) to $0.890$ (for soybean).
Additionally, compared to its in-season prediction results
 (see the $6$th row of Table~\ref{tab:exp-crop-pred}), 
the one year ahead prediction outcomes of our approach are just slightly inferior,
degraded respectively by $8.6 \%$, of $15.5 \%$, of $17.0 \%$, and of $14.6 \%$ 
in terms of RMSE for corn, cotton, soybean, and winter wheat.
The empirical results confirm the robustness of MMST-ViT.

Interestingly, we also observe that 
the prediction results for the cotton have a high RMSE value but not a low R$^2$ (or Corr) value.
This is because the cotton yield is measured by pounds per acre (LB/AC), 
with a high standard deviation value of $250.1$,
while other crop yields are measured by bushels per acre (BU/AC),
with low standard deviation values (\eg, $11.8$ for soybean).
In scenarios with high variance data, the RMSE value may worsen due to larger residuals, 
while the R$^{2}$ value improves due to the increased proportion of explained variance.

\subsection{Ablation Studies}
\label{sec:exp-alation-study}

%
\begin{table}  [!t] 
    \scriptsize
    \centering
    \setlength\tabcolsep{4pt}
    \caption{
        Ablation studies for different components,
        with five scenarios considered and
        the last row listing the results of MMST-ViT
        }
        \vspace{-0.5 em}
        \begin{tabular}{@{}|cc|cc|cc|@{}}
            \toprule
            \multicolumn{2}{|c|}{\multirow{2}{*}{Component}} & \multicolumn{2}{c|}{Corn}         & \multicolumn{2}{c|}{Soybean}      \\ \cmidrule(l){3-6} 
            \multicolumn{2}{|c|}{}                           & \multicolumn{1}{c}{RMSE ($\downarrow$)} & Corr ($\uparrow$)  & \multicolumn{1}{c}{RMSE ($\downarrow$)} & Corr ($\uparrow$)  \\ \midrule
            \multicolumn{1}{|c|}{\multirow{2}{*}{\begin{tabular}[c]{@{}c@{}}Temporal\\ Transformer\end{tabular}}} &
              w/o long-term &
              \multicolumn{1}{c}{14.5} &
              0.843 &
              \multicolumn{1}{c}{6.2} &
              0.850 \\ 
            \multicolumn{1}{|c|}{}     & w/o T-MHA           & \multicolumn{1}{c}{15.7} & 0.820  & \multicolumn{1}{c}{5.6}  & 0.839 \\ \midrule
            \multicolumn{1}{|c|}{\begin{tabular}[c]{@{}c@{}}Spatial \\ Transformer\end{tabular}} &
              w/o S-MHA &
              \multicolumn{1}{c}{13.5} &
              0.856 &
              \multicolumn{1}{c}{5.6} &
              0.849 \\ \midrule
            \multicolumn{1}{|c|}{\multirow{2}{*}{\begin{tabular}[c]{@{}c@{}}Multi-Modal \\ Transformer\end{tabular}}} &
              w/o short-term &
              \multicolumn{1}{c}{13.6} &
              0.839 &
              \multicolumn{1}{c}{6.1} &
              0.845 \\ 
            \multicolumn{1}{|c|}{}     & w/o image & \multicolumn{1}{c}{15.2} & 0.809 & \multicolumn{1}{c}{6.8}  & 0.822 \\ \midrule
            \multicolumn{1}{|c|}{MMST-ViT} & -                   & \multicolumn{1}{c}{10.5} & 0.900   & \multicolumn{1}{c}{3.9}  & 0.918 \\ \bottomrule
            \end{tabular}
    
    \label{tab:exp-as-component}
    \vspace{-1.0 em}
    
\end{table}

\par\smallskip\noindent
{\bf Key Components.} 
We next conduct experiments to show  how each key component 
in our MMST-ViT affects prediction performance.
Table~\ref{tab:exp-as-component} lists the results of our ablation studies under five different scenarios,
with one shown in a row.
Here, ``w/o long-term'' indicates masking the long-term meteorological data,
and ``w/o T-MHA'' represents the absence of our Temporal Transformer 
and instead resorting to the average pooling to obtain the global temporal representation.
Similarly, ``w/o S-MHA'' denotes utilizing the average pooling to obtain the global spatial representation.
The scenarios of ``w/o short-term'' and ``w/o image'' represent 
respectively masking the short-term meteorological data and the remote sensing data from the Multi-Modal Transformer.
Note that masking either data makes our model unable to conduct MM-MHA, following \eqref{eq:mm-attn}.
The results of our complete MMST-ViT outcomes are shown in the last row of Table~\ref{tab:exp-as-component}.

Three observations can be made from Table~\ref{tab:exp-as-component}.
First, the long-term meteorological data degrades the Corr value
by $0.057$ (or $0.068$) for corn (or soybean).
This validates that crop yield prediction accuracy can benefit from historical meteorological data,
which dictates long-term climate change.
Second, the absence of either the Temporal Transformer or the Spatial Transformer
lowers the Corr value by $0.080$ (for corn) and by $0.069$ (for soybean), respectively.
Third, unable to conduct MM-MHA as the result of masking
the short-term meteorological data or the satellite images
degrades prediction performance greatly.
For example, masking the short-term meteorological data (or satellite images)
causes the Corr value degradation by $0.096$ (or $0.073$) for soybean.
This statistical evidence exhibits the importance of our Multi-Modal Transformer 
for capturing the impact of meteorological parameters on crop growth.

\par\smallskip\noindent
{\bf Pre-training Techniques.} 
We conduct experiments to explore the impact of our proposed multi-modal self-supervised pre-training,
\ie, \eqref{eq:loss-sim-clr}, on the crop yield prediction.
We consider three scenarios,
\ie, MMST-ViT without pre-training, 
with the pre-training technique described in SimCLR~\cite{chen:icml20:simclr},
and with our multi-modal pre-training.
Table~\ref{tab:exp-as-pre-train} presents the prediction performance outcomes.
The table reveals that our MMST-ViT (w/ our multi-modal pre-training) significantly outperforms
its counterpart (w/o pre-training),
exhibiting a lower RMSE value by $2.7$ (or $1.2$) 
and a larger Corr value by $0.043$ (or $0.043$) for corn (or soybean),
due to two reasons.
First, Vision Transformer (ViT)-based models are prone to overfitting~\cite{dosovitskiy:iclr21:vit},
but our multi-modal self-supervised learning can significantly mitigate this issue 
by leveraging both visual remote sensing data and numerical meteorological data for pre-training.
Second, our multi-modal self-supervised pre-training may can capture the impact of meteorological parameters
on the crops.
In contrast,
pre-training our MMST-ViT with the SimCLR technique 
only achieves marginal performance improvement (compared to the one w/o pre-training),
achieving the Corr improvement of $0.001$ ($0.875$ v.s. $0.876$) for soybean.
This is because the SimCLR technique fails to leverage numerical meteorological parameters for pre-training.
These empirical results validate the necessity and importance 
of our proposed multi-modal pre-training for accurate crop yield predictions.

%
\begin{table} [!t]
    \scriptsize
    \centering
    \setlength\tabcolsep{5 pt}
    \caption{
        Ablation studies for different pre-training techniques, with three scenarios considered
        }
        \vspace{-0.5 em}
        \begin{tabular}{@{}|c|cc|cc|@{}}
            \toprule
            \multirow{2}{*}{Method} & \multicolumn{2}{c|}{Corn}         & \multicolumn{2}{c|}{Soybean}      \\ \cmidrule(l){2-5} 
                                    & \multicolumn{1}{c}{RMSE ($\downarrow$)} & Corr ($\uparrow$)  & \multicolumn{1}{c}{RMSE ($\downarrow$)} & Corr ($\uparrow$)  \\ \midrule
            w/o pretraining         & \multicolumn{1}{c}{13.2} & 0.857 & \multicolumn{1}{c}{5.1}  & 0.875 \\ 
            w/ SimCLR               & \multicolumn{1}{c}{12.9} & 0.857 & \multicolumn{1}{c}{4.9}  & 0.876 \\
            w/ multi-modal pretraining          & \multicolumn{1}{c}{10.5 } & 0.900 & \multicolumn{1}{c}{3.9}  & 0.918 \\ \bottomrule
            \end{tabular}
    
    \label{tab:exp-as-pre-train}
    \vspace{-1.5 em}
\end{table}

\section{Conclusion}\label{sec:conslusion}
This paper has proposed the Multi-Modal Spatial-Temporal Vision Transformer (MMST-ViT),
a climate change-aware deep learning approach for predicting crop yields at the county level 
across the United States.
MMST-ViT comprises three key components of a Multi-Modal Transformer, a Spatial Transformer, and a Temporal Transformer. 
Three innovative Multi-Head Attention (MHA) mechanisms are introduced, one for each component.
As a result, our MMST-ViT can leverage both the visual remote sensing data
and the numerical meteorological data 
for capturing the impact of short-term growing season weather variations
and long-term climate change on crop yields.
Additionally, a novel multi-modal contrastive learning technique has been developed,
able to effectively pre-train our model 
without the need of human supervision.
We have conducted extensive experiments on $200$+ counties/parishes located in $4$ U.S. states,
with the results demonstrating that our proposed MMST-ViT 
substantially outperforms its state-of-the-art counterparts consistently under three performance metrics of interest.

\section*{Acknowledgments}
This work was supported in part by NSF under Grants 2019511, 2146447, 
and in part by the BoRSF under Grants LEQSF(2019-22)-RD-A-21 and LEQSF(2021-22)-RD-D-07. 
Any opinion and findings expressed in the paper are those of the authors 
and do not necessarily reflect the view of funding agencies.

{\small
\bibliographystyle{ieee_fullname}
\bibliography{egbib}
}

\end{document}